\documentclass[10pt,twocolumn,letterpaper]{article}

\usepackage[pagenumbers]{cvpr} 

\usepackage{graphicx}
\usepackage{amsmath}
\usepackage{amssymb}
\usepackage{booktabs}
\usepackage{multirow}
\usepackage{appendix}
\usepackage{stfloats}
\usepackage[pagebackref,breaklinks,colorlinks]{hyperref}
\usepackage[capitalize]{cleveref}
\crefname{section}{Sec.}{Secs.}
\Crefname{section}{Section}{Sections}
\Crefname{table}{Table}{Tables}
\crefname{table}{Tab.}{Tabs.}

\def\cvprPaperID{2766} 
\def\confName{CVPR}
\def\confYear{2023}

\begin{document}

\title{SOOD: Towards Semi-Supervised Oriented Object Detection}

\author{Wei Hua$^{*1}$, Dingkang Liang$^{*1}$, Jingyu Li$^{1}$, Xiaolong Liu$^{1}$, Zhikang Zou$^{2}$, Xiaoqing Ye$^{2}$, Xiang Bai$^{\dag1}$\\
\\
        $^{1}$Huazhong University of Science and Technology, \{whua\_hust, dkliang, xbai\}@hust.edu.cn\\
        $^{2}$ Baidu Inc., China\\
}

\maketitle
\protect \renewcommand{\thefootnote}{\fnsymbol{footnote}}
\footnotetext[1]{Equal contribution. $^{\dag}$Corresponding author.} 
\footnotetext[0]{Work done when Dingkang Liang was an intern at Baidu.}

\maketitle

\begin{abstract}

Semi-Supervised Object Detection~(SSOD), aiming to explore unlabeled data for boosting object detectors, has become an active task in recent years. However, existing SSOD approaches mainly focus on horizontal objects, leaving multi-oriented objects that are common in aerial images unexplored. This paper proposes a novel Semi-supervised Oriented Object Detection model, termed SOOD, built upon the mainstream pseudo-labeling framework. Towards oriented objects in aerial scenes, we design two loss functions to provide better supervision. Focusing on the orientations of objects, the first loss regularizes the consistency between each pseudo-label-prediction pair (includes a prediction and its corresponding pseudo label) with adaptive weights based on their orientation gap. Focusing on the layout of an image, the second loss regularizes the similarity and explicitly builds the many-to-many relation between the sets of pseudo-labels and predictions. Such a global consistency constraint can further boost semi-supervised learning. Our experiments show that when trained with the two proposed losses, SOOD surpasses the state-of-the-art SSOD methods under various settings on the DOTA-v1.5 benchmark. The code will be available at \url{https://github.com/HamPerdredes/SOOD}.

\end{abstract}

\section{Introduction}
\label{sec:intro}

Sufficient labeled data is essential for fully-supervised object detection. However, the data labeling process is time-consuming and expensive. Recently, Semi-Supervised Object Detection~(SSOD), where object detectors are learned from labeled data as well as easy-to-obtain unlabeled data, has attracted increasing attention. Existing SSOD methods~\cite{liu2021unbiased,xu2021end,zhou2022dense,li2022pseco} mainly focus on detecting objects with horizontal bounding boxes in general scenes. Nevertheless, in more complex scenes, such as aerial scenes, objects usually need to be annotated with oriented bounding boxes. Considering the higher annotation cost of oriented boxes\footnote{Annotation cost of an oriented box is about 36.5\%~(86\$ vs. 63\$ per 1k at 2022.11) more than a horizontal box according to \url{https://cloud.google.com/ai-platform/data-labeling/pricing}.}, semi-supervised oriented object detection is worth studying.

\begin{figure}[t]
	\begin{center}
		\includegraphics[width=0.96\linewidth]{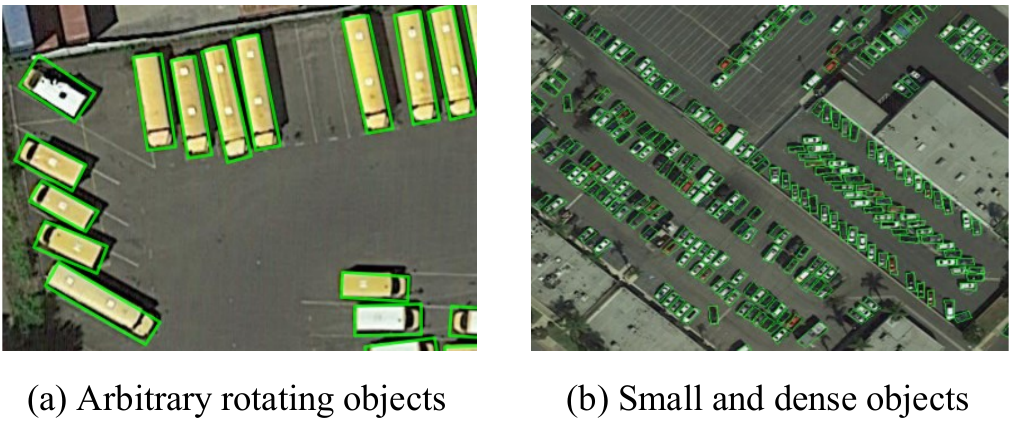}
	\end{center}
	\caption{Arbitrary rotating (a), small and dense (b) objects are common in aerial scenes, which are often regularly arranged on the image. From a global perspective, this pattern indicates that an aerial can be regarded as a layout.}
	\label{fig:intro1}
\end{figure}

Compared with general scenes, the main characteristics of objects in aerial scenes~(or aerial objects for short) are three-fold: arbitrary orientations, small scales, and agglomeration, as shown in Fig.~\ref{fig:intro1}. The mainstream SSOD methods are based on the pseudo-labeling framework~\cite{chen2022dense,tarvainen2017mean,tang2021humble} consisting of a teacher model and a student model. The teacher model, an Exponential Moving Average (EMA) of the student model at historical training iterations, generates pseudo-labels for unlabeled images. Thus, the student model can learn from both labeled and unlabeled data.
To extend the framework to oriented object detection, we think the following two aspects need to be addressed: 1) As orientation is an essential property of multi-oriented objects, how to use the orientation information when guiding the student with pseudo-labels is critical. 2) As aerial objects are often dense and regularly distributed in an image, we can utilize the layout to facilitate the learning of each pair instead of treating them individually.

This paper proposes the first Semi-supervised Oriented Object Detection method, termed SOOD. Following \cite{zhou2022dense}, SOOD is built upon the dense pseudo-labeling framework, where the pseudo labels are filtered from the raw pixel-wise predictions~(including box coordinates and confidence scores).
The key design is two simple yet effective losses that enforce the instance-level and set-level consistency between the student's and the teacher's predictions.

To be specific, considering that the pseudo-label-prediction pairs are not equally informative, we propose the Rotation-aware Adaptive Weighting~(RAW) loss. It utilizes the orientation gap of each pair, which reflects the difficulty of this sample in a way, to weight the corresponding loss dynamically. In this manner, we can softly pick those more useful supervision signals to guide the learning of the student. 
In addition, considering that the layout of an aerial image can potentially reflect components’ overall status (e.g., objects’ density and location distribution) and help the detection process, we propose the Global Consistency (GC) loss. It measures the similarity of the pseudo-labels and the predictions from a global perspective, which can alleviate the disturbance of noise in pseudo-labels and implicitly regularizes the mutual relations between different objects.

We extensively evaluate SOOD under various settings on DOTA-v1.5, a popular aerial object detection benchmark. Our SOOD achieves consistent performance improvement when using 10\%, 20\%, 30\%, and full of labeled data, compared with the state-of-the-art SSOD methods (using the same oriented object detector). The ablation study also verifies the effectiveness of the two losses. 

In summary, this paper makes an early exploration of semi-supervised learning for oriented object detection. By analyzing the distinct characteristics of oriented objects from general objects, we propose two novel loss functions to adapt the pseudo-label framework to this task. We hope that this work can provide a good starting point for semi-supervised oriented object detection and serve as a simple yet strong baseline for future research.

\section{Related works}

\noindent\textbf{Semi-Supervised Object Detection.}
In the past few years, semi-supervised learning~(SSL)~\cite{sohn2020fixmatch,berthelot2019mixmatch} has achieved impressive performance in image classification. These works leverage unlabeled data by using pseudo-label~\cite{lee2013pseudo,grandvalet2004semi,xie2020self,li2023dds3d}, consistency regularization~\cite{tarvainen2017mean,xie2020unsupervised,berthelot2019mixmatch}, data augmentation~\cite{sajjadi2016regularization,chen2020simple} and even adversarial training~\cite{miyato2018virtual}. Compared to semi-supervised image classification, SSOD requires instance-level predictions and additional bounding boxes regression sub-task, which makes it more challenging. 
In~\cite{radosavovic2018data,zoph2020rethinking}, pseudo-labels are assembled from predictions of data with different augmentations. CSD~\cite{jeong2019consistency} only utilizes the horizontal flipping augmentation and applies consistency loss to constrain the model, but the weak augmentation limits its performance. STAC~\cite{sohn2020simple} trains an object detection with labeled data and generates pseudo-labels on unlabeled data with this detector offline. After that, some studies~\cite{liu2021unbiased,tang2021humble,xu2021end,yang2021interactive} adopt EMA from Mean Teacher~\cite{tarvainen2017mean} to update the teacher model after each training iteration. ISMT~\cite{yang2021interactive} obtains more accurate pseudo-labels by fusing current pseudo-labels and history labels. Unbiased Teacher~\cite{liu2021unbiased} replaces cross-entropy loss with focal loss~\cite{lin2017focal} to solve the class-imbalance issue and filters pseudo-labels by threshold. Soft Teacher~\cite{xu2021end} uses the classification scores to adaptively weight the loss of each pseudo-box and proposes box jittering to select reliable pseudo-labels. Unbiased Teacher v2~\cite{liu2022unbiased} 
adopts an anchor-free detector and uses uncertainty predictions to select pseudo-labels for the regression branch. Dense Teacher~\cite{zhou2022dense} replaces post-processed instance-level pseudo-labels with dense pixel-level pseudo-labels, which successfully removes the influence of thresholds and post-processing hyper-parameters. However, none of these works are designed for oriented object detection in aerial scenes. This paper aims to fill this blank and offer a starting point for future research.

\noindent\textbf{Orient Object Detection.}
Different from general object detectors~\cite{girshick2015fast,ren2015faster,liu2016ssd,redmon2016you}, oriented object detectors represent objects with Oriented Bounding Boxes~(OBBs). Typical oriented objects include aerial objects and multi-oriented scene texts~\cite{liao2020real,tang2022few,liao2018rotation,han2021redet}. In recent years, many oriented object detection methods have been proposed to boost the performance for this area. CSL~\cite{yang2022arbitrary} formulates the angle regression problem as a classification task to address the out-of-bound issue. R$^3$Det~\cite{yang2021r3det} predicts Horizontal Bounding Boxes~(HBBs) at the first stage to improve detection speed and align the feature in the second stage to predict oriented objects. 
Oriented R-CNN~\cite{xie2021oriented} proposes a concise multi-oriented region proposal network and uses the midpoint offsets to represent arbitrarily oriented objects. ReDet~\cite{han2021redet} proposes a Rotation-equivariant detector to extract rotation-invariant from rotation-equivariant for accurate aerial object detection. Oriented RepPoints~\cite{li2022oriented} proposes a quality assessment module and samples assignment scheme for adaptive points learning, which can obtain non-axis features from neighboring objects and neglect background noises. Different from the above works that focus on the supervised paradigm, this paper makes an early exploration of semi-supervised oriented object detection, which can reduce the annotation cost and boost detectors with unlabeled data.

\section{Preliminary}
In this section, we revisit the mainstream pseudo-labeling paradigm in SSOD and Monge-Kantarovich optimal transport theory~\cite{monge1781memoire} as preliminary.

\subsection{Pseudo-labeling Paradigm}
\label{src:preliminary_pl}
Pseudo-labeling frameworks~\cite{xu2021end, liu2022unbiased, zhou2022dense} inherited designs from the Mean Teacher~\cite{tarvainen2017mean} structure, which consists of two parts, i.e., a teacher model and a student model. The teacher model is an Exponential Moving Average~(EMA) of the student model. They are learned iteratively by the following steps.
1) Generate pseudo-labels for the unlabeled data in a batch. The pseudo-labels are filtered from the teacher's predictions, e.g., the box coordinates and the classification scores. Meanwhile, the student makes predictions for both labeled and unlabeled data in the batch.
2) Compute loss for the student model's predictions. It consists of two parts, the unsupervised loss $\mathcal{L}_u$ and supervised loss $\mathcal{L}_s$. They are computed for the unlabeled data with the pseudo-labels and the labeled data with the ground truth (GT) labels, respectively.
The overall loss $\mathcal{L}$ is the sum of them.
3) Update the parameters of the student model according to the overall loss. The teacher model is updated simultaneously in an EMA manner. In this way, based on the mutual learning mechanism, both models evolve as the training goes on.

Based on the sparsity of pseudo-labels, pseudo-labeling frameworks can be further categorized into sparse pseudo-labeling ~\cite{xu2021end, liu2022unbiased} and dense pseudo-labeling ~\cite{zhou2022dense}, termed SPL and DPL, respectively. The SPL selects the teacher's predictions after the post-processing operations, e.g., non-maximum suppression and score filtering. It obtains sparse labels to supervise the student, e.g., bounding boxes and categories. The DPL directly samples the post-sigmoid logits predicted by the teacher, which are dense and informative. Compared with SPL, DPL bypasses those lengthy post-processing methods, reserving more details from the teacher than its pseudo-box counterpart.

\subsection{Optimal Transport}
\label{src:preliminary_ot}
The Monge-Kantorovich Optimal Transport (OT)~\cite{monge1781memoire} aims to solve the problem of simultaneously moving items from one set to another set with minimum cost. It has been widely explored in various computer vision tasks~\cite{ge2021ota, arjovsky2017wasserstein, wang2020distribution, zhan2021unbalanced}. The mathematical formulations of OT are described as follows in detail.

Let $\mathbb{X} = \{x_i | x_i \in \mathbb{R}^d\}_i^N$ and $\mathbb{Y} = \{y_j | y_j \in \mathbb{R}^d\}_j^N$ denote two sets of $N$ $d$-dimensional vectors. Their discrete distributions $\hat{\mathbb{X}}$ and $\hat{\mathbb{Y}}$ are formulated as:
\begin{equation}
    \hat{\mathbb{X}} = {\textstyle \sum_{i}^{N}} \hat{x}_i\delta _{f_i} 
\end{equation}
\begin{equation}
     \hat{\mathbb{Y}} = {\textstyle \sum_{j}^{N}} \hat{y}_j\delta _{g_j},
\end{equation}
where $\mathbf{\hat{x}}$ and $\mathbf{\hat{y}}$ are the discrete probability vectors, $\delta$ is the Dirac delta function. Therefore, the OT cost is measured between these two probabilities, $\mathbf{\hat{x}}$ and $\mathbf{\hat{y}}$. The possible transportation plans from $\mathbf{\hat{x}}$ to $\mathbf{\hat{y}}$ are formed as: 
\begin{equation}
    \boldsymbol{P} = \{\boldsymbol{p}~\in\mathbb{R}^{N\times N} | \boldsymbol{p}\mathbf{1}_N = \mathbf{\hat{x}}, \boldsymbol{p}^T\mathbf{1}_N = \mathbf{\hat{y}}\},
\end{equation}
where $\mathbf{1}_N$ is an $N$-dimensional column vector whose values are all 1. The OT cost is then defined as: 
\begin{equation}
    \omega_{ot}(\mathbf{\hat{x}}, \mathbf{\hat{y}}) = \min_{\boldsymbol{p}\in\boldsymbol{P}}\left \langle \mathbf{C}, \boldsymbol{p} \right \rangle,
\end{equation}
where $\mathbf{C}\in \mathbb{R}^{N\times N}$ represents the cost matrix between two sets, $\left \langle \cdot \right \rangle$ represents inner product. In common, the OT problem is solved in its dual formulation
\begin{equation}
\begin{split}
    \mathcal{W}_{ot}(\mathbf{\hat{x}}, \mathbf{\hat{y}}) = \max_{\boldsymbol{\lambda},\boldsymbol{\mu}\in\mathbb{R}^N}
    \left \langle \boldsymbol{\lambda}, \mathbf{\hat{x}}  \right \rangle + 
    \left \langle \boldsymbol{\mu}, \mathbf{\hat{y}}  \right \rangle,\\
    s.t.~~\lambda_i + \mu_j \le \mathbf{C}_{i, j},~\forall i, j,
\end{split}
\end{equation}
where $\boldsymbol{\lambda}$ and $\boldsymbol{\mu}$ are the solutions of the OT problem, which can be approximated in an iterative manner~\cite{cuturi2013sinkhorn}.

\section{Method}
\begin{figure*}[t]
	\begin{center}
		\includegraphics[width=0.96\linewidth]{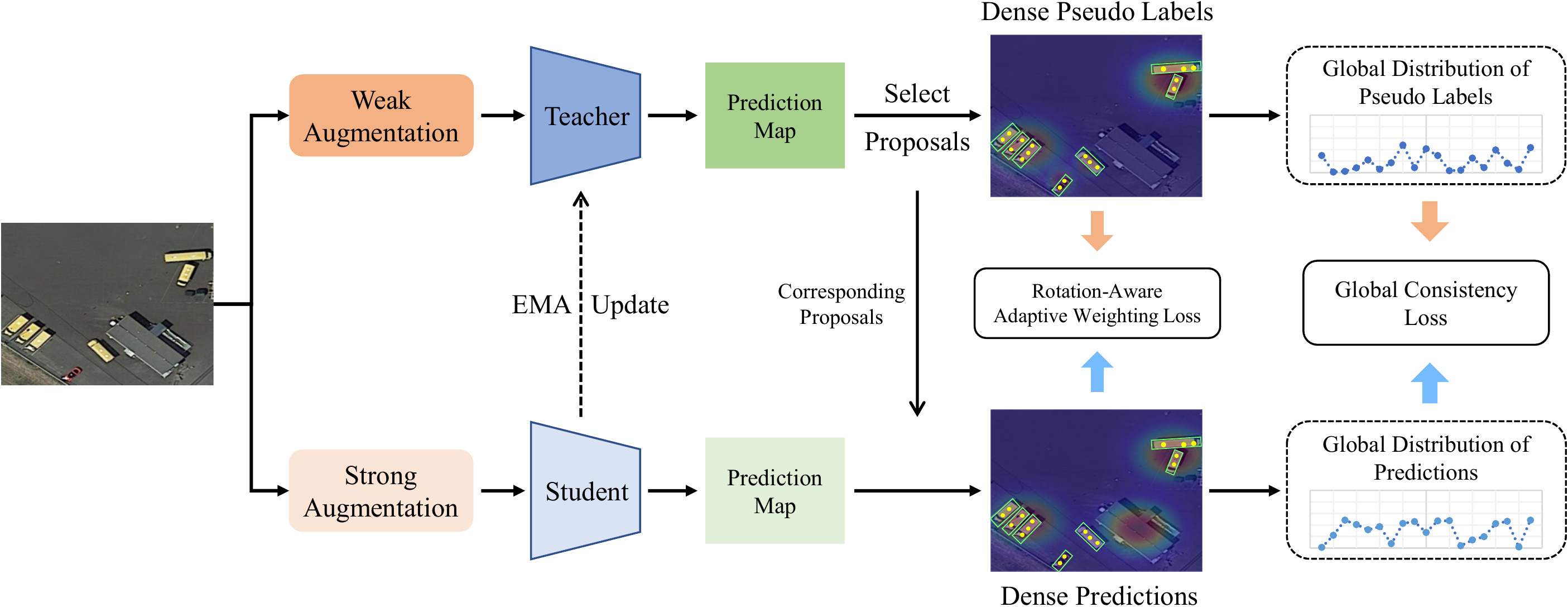}
	\end{center}
	\caption{The pipeline of the proposed SOOD. Each training batch consists of both labeled and unlabeled images. Note that the supervised part is hidden for simplicity. For the unsupervised part, we sample dense pseudo-labels from the teacher's prediction map and select the student's predictions at the same positions, obtaining a series of pseudo-label-prediction pairs. We dynamically weigh each pair's unsupervised loss by their orientation difference. Besides, we regard them as two discrete distributions, measuring their similarity in a many-to-many manner via a global consistency loss.}
	\label{fig:pipeline}
\end{figure*}

Fig.~\ref{fig:pipeline} illustrates an overview of our proposed SOOD. Towards multi-oriented object detection in aerial images, we build our approach upon the popular dense pseudo-labeling framework~\cite{zhou2022dense}, along with the Rotation-aware Adaptive Weighting (RAW) loss and the Global Consistency (GC) loss. In this section, we first describe the overall framework in Sec.~\ref{sec:overall}. Then, we describe the key design of the proposed losses, RAW and GC, in the following Sec.~\ref{sec:loss1} and Sec.~\ref{sec:loss2}, respectively.

\subsection{The Overall Framework}
\label{sec:overall}
Currently, the Dense Pseudo-Labeling (DPL) framework achieves the state-of-the-art in SSOD. Hence, we construct a DPL-based end-to-end baseline, including the supervised and unsupervised parts. For the supervised part, the student model is trained with labeled data in a regular manner. For the unsupervised part, we first obtain predicted boxes of the teacher after post-processing. These boxes indicate informative areas in the prediction map, where we randomly sample predictions, forming them as dense pseudo-labels $P^t$. Note that we also select the predictions $P^s$ at the same correspondence positions from the student.

We use the oriented version of FCOS~\cite{tian2019fcos} as the teacher and student models. The basic unsupervised loss consists of three parts: classification loss, regression loss, and center-ness loss, corresponding to the output of FCOS. We adopt smooth l1 loss for regression loss,
binary cross-entropy loss for classification and center-ness loss. Based on these losses, we first perform adaptive weighting on them through RAW and further measure the global consistency between the teacher and the student via GC.

\subsection{Rotation-aware Adaptive Weighting Loss}
\label{sec:loss1}
Orientation is one essential property of an oriented object. As shown in Fig.~\ref{fig:intro1}, even if objects are dense and small, their orientations remain clear. Previous oriented object detection methods have already employed such a property by assembling it into loss calculation. However, these works are under the assumption that the angles of the labels are reliable. In this case, it is natural to strictly force the prediction close to the ground truth.

Unfortunately, the above assumption does not hold in the semi-supervised setting. In other words, the pseudo-labels may be incorrect. Simply forcing the student to be close to the teacher may cause noise accumulation, harming the model's training process. Hence, we propose to utilize the orientation information softly. Intuitively, as orientation is essential but hard to be accurately predicted, the difference in rotation angles between a prediction and a pseudo-label can reflect the difficulty of the sample in a way. In other words, the orientation difference can be used to dynamically adjust the unsupervised loss. Therefore, we construct a rotation-aware modulating factor, similar to focal loss~\cite{lin2017focal}. This factor can dynamically weight the loss of each pseudo-label-prediction pair by considering their orientation difference.

Specifically, the modulating factor $\omega_i^{rot}$ of the $i$-th pair is formed as:
\begin{equation}
\label{eq6}
\omega_i^{rot} = 1 + \sigma_i,
\end{equation}
\begin{equation}
\label{eq7}
\sigma_i = \alpha \frac{\left | r_i^t - r_i^s \right | }{\pi},  r_i^t, r _i^s \in [-\frac{\pi}{2}, \frac{\pi}{2}),
\end{equation}
where $r_i^t$ and $r_i^s$ are the $i$-th pseudo-label's and prediction's rotation angle in radians, respectively. $\alpha$ is a hyper-parameter for adjusting orientation's importance, and we set it to 50 empirically. We add a constant to $\sigma_i$, maintaining the origin unsupervised loss when pseudo-label and prediction have the same orientation. With the rotation-aware modulating factor, the overall rotation-aware adaptive weighting loss is formulated as:
\begin{equation}
\mathcal{L}_{RAW} = {\textstyle \sum_{i}^{N_p}} \omega_i^{rot} \mathcal{L}_u^i,
\end{equation}
where $N_p$ is the number of pseudo-labels and $\mathcal{L}_u^i$ is the basic unsupervised loss of the $i$-th pseudo-label-prediction pair. By using the rotation-aware modulating factor, the RAW loss makes better use of the orientation information and provides more informative guidance, potentially benefiting the semi-supervised learning process.

\subsection{Global Consistency Loss}
\label{sec:loss2}

Objects in an aerial image are usually dense and regularly distributed, as depicted in Fig.~\ref{fig:intro1}. Similar to texts in a document, the arrangement of the set of objects, i.e., the layout, encodes the mutual relations between them and the global pattern of the image.
Ideally, the layout consistency between the student's and the teacher's predictions will be ensured if each pseudo-label-prediction pair is aligned. However, the latter condition is too strict and may hurt performance when there are noises in pseudo-labels. Therefore, it is reasonable to add the consistency between layouts as an additional \textit{relaxed} optimization objective, encouraging the student to learn robust information from the teacher.

In this way, the noise disturbance in pseudo-labels can be alleviated. Besides, the relations between different predicted instances from the student can also be regularized implicitly, which provides an additional guide to the student.

We introduce the optimal transport cost~\cite{villani2009optimal} to measure the global similarity of layouts between the teacher's and the student's predictions, forming the global consistency loss. To be concrete, we denote the classification scores predicted by the teacher and the student by $\mathbf{s}^t\in\mathbb{R}^{N_p\times K}$ and $\mathbf{s}^s\in\mathbb{R}^{N_p\times K}$ respectively, where $K$ is the number of classes. Then, their global distributions, $\mathbf{d}^t~\in~\mathbb{R}^{N_p}$ and $\mathbf{d}^s~\in~\mathbb{R}^{N_p}$ can be formulated by
\begin{equation}
    \mathbf{d}_i^t = e^{\mathbf{s}_{i, c(i)}^t},
\end{equation}
\begin{equation}
    \mathbf{d}_i^s = e^{\mathbf{s}_{i, c(i)}^s},
\end{equation}
where $c(i)=\mathop{\arg\max}\limits_{j=1, ..., K} \mathbf{s}_{i, j}^t$ is the index of the class with the largest score for the $i$-th pseudo-label.

The global consistency loss is defined as the OT problem's dual formulation
\begin{equation}
\begin{aligned}
   \mathcal{L}_{G C}(\mathbf{d}^t, \mathbf{d}^s) & = \left\langle\boldsymbol{\lambda}^{*}, 
   \frac{\mathbf{d}^t}{\|\mathbf{d}^t\|_1}\right\rangle+\left\langle\boldsymbol{\mu}^{*},
   \frac{\mathbf{d}^s}{\|\mathbf{d}^s\|_1}\right\rangle,
\end{aligned}
\end{equation}
where we normalize two distributions to form discrete probabilities, by dividing them to their sum. To construct the cost map for solving the OT problem, we consider both the spatial distance and the score difference of each possible matching pair. Specifically, for each prediction, we measure its matching cost with every pseudo-label as follows:
\begin{equation}
    C_{i, j} = C_{i, j}^{dist} + C_{i, j}^{score},
\end{equation}
\begin{equation}
    C_{i, j}^{dist} = \frac{\|\mathbf{z}_i^t - \mathbf{z}_j^s\|_2^2}{\max_{1<=a, b<=N_p} \|\mathbf{z}_a^t - \mathbf{z}_b^s\|_2^2},
\end{equation}
\begin{equation}
    C_{i, j}^{score} = \frac{\|\mathbf{s}_{i, c(i)}^t - \mathbf{s}_{j, c(j)}^s\|_1}{\max_{1<=a, b<=N_p}
    \|\mathbf{s}_{a, c(a)}^t - \mathbf{s}_{b, c(b)}^s\|_1},
\end{equation}
where $\mathbf{z}_i^t$ and $\mathbf{z}_j^s$ are 2D coordinates of the $i$-th sample in the teacher and the $j_{th}$ sample in the student.

We solve the OT problem by a fast Sinkhorn distances algorithm~\cite{cuturi2013sinkhorn}, obtaining the approximate solution $\boldsymbol{\lambda}^{*}$ and $\boldsymbol{\mu}^{*}$. Based on the defined loss, its gradient with respect to $\mathbf{d}^s$ is
\begin{equation}
\frac{\partial \mathcal{L}_{G C}(\mathbf{d}^t, \mathbf{d}^s)}{\partial \mathbf{d}^s}=\frac{\boldsymbol{\mu}^{*}}{\|\mathbf{d}^s\|_{1}}-\frac{\left \langle\boldsymbol{\mu}^{*}, \mathbf{d}^s\right \rangle}{\|\mathbf{d}^s\|_{1}^{2}} .
\end{equation}

The gradients can be back-propagated to update the model, enforcing the layout consistency in the framework. Although OT-based loss has been explored before~\cite{wang2020distribution,zhan2021unbalanced,frogner2015learning}, our goal of using OT is different. Specifically, they focus on utilizing OT to improve the model's generalizability~\cite{wang2020distribution,frogner2015learning} or mitigate the matching constraint~\cite{zhan2021unbalanced}. However, our GC aims to model the many-to-many relationship between the teacher and the student, which is complementary to the RAW. In addition, we adopt such a set-to-set matching to alleviate the error in pseudo-label assignment, providing a more loose but stable constraint.

SOOD is trained with the proposed unsupervised losses, RAW and GC, for unlabeled data as well as the supervised loss for labeled data. The overall loss 
$\mathcal{L}$ is defined as:
\begin{equation}
    \mathcal{L} =  \underbrace{\mathcal{L}_{RAW} + \mathcal{L}_{GC}}_{\mathcal{L}_u} + \mathcal{L}_s.
\end{equation}

Note that the supervised loss is the same as defined in FCOS, our designs only modify the unsupervised part. 

\section{Experiments}

We conduct experiments on DOTA-v1.5, which is proposed at DOAI-2019\footnote{The 1st Workshop on Detecting Objects in Aerial Images in conjunction with IEEE CVPR 2019~\url{https://captain-whu.github.io/DOAI2019/dataset.html}}. It contains 2806 large aerial images and 402,089 annotated oriented objects. It includes three subsets: DOTA-v1.5-train, DOTA-v1.5-val and DOTA-v1.5-test, containing 1411, 458, and 937 images, respectively. The annotations of DOTA-v1.5-test is not released.

There are 16 categories in this dataset: Plane~(PL), Baseball diamond~(BD), Bridge~(BR), Ground track field~(GTF), Small vehicle~(SV), Large vehicle~(LV), Ship~(SH), Tennis court~(TC), Basketball court~(BC), Storage tank~(ST), Soccer-ball field~(SBF), Roundabout~(RA), Harbor~(HA), Swimming pool~(SP), Helicopter~(HC) and Container crane~(CC). Compared with DOTA-v1.0~\cite{xia2018dota}, a previous version, DOTA-v1.5 contains more small instances~(less than 10 pixels), which makes it more challenging.

Following conventions in SSOD, we consider two protocols, Partially Labeled Data and Fully Labeled Data, to validate the performance of a method on limited and abundant labeled data, respectively.

\noindent\textbf{Partially Labeled Data.} We randomly sample 10\%, 20\%, and 30\% images from DOTA-v1.5-train as labeled data and set the remaining images as unlabeled data. Following DOTA-v1.5-train's data distribution, we provide one fold for each data proportion.

\noindent\textbf{Fully Labeled Data.} We set DOTA-v1.5-train as labeled data and DOTA-v1.5-test as unlabeled data.

For all experiments, we perform evaluation on DOTA-v1.5-val and report the performance with the standard mean average precision~(mAP) as the evaluation metrics.

\begin{table*}[t]
\small
\centering
\setlength{\tabcolsep}{6.0mm}
\caption{Experimental results on DOTA-v1.5 under the Partially Labeled Data setting. * and \dag~indicate our implementations with rotated-Faster-RCNN and rotated-FCOS, respectively. Experiments are conducted on 10\%, 20\% and 30\% labeled data settings.}
\label{tab:experiments}
\begin{tabular}{cccccc}
\toprule
Setting&Method &Publication&10\% &20\%&30\% \\
\midrule
{\multirow{2}{*}{Supervised}}&Faster R-CNN*~\cite{ren2015faster}& NeurIPS 2016& 43.43 & 51.32 & 53.14 \\
&FCOS\dag~\cite{tian2019fcos}&ICCV 2019 &42.78 &50.11 &54.79 \\
\midrule
{\multirow{5}{*}{Semi-supervised}}&Unbiased Teacher*~\cite{liu2021unbiased} & ICLR 2021 &44.51 &52.80 &53.33 \\
&Soft Teacher*~\cite{xu2021end} & ICCV 2021 &48.46 &54.89 &57.83 \\
&Dense Teacher\dag~\cite{zhou2022dense} & ECCV 2022 & 46.90 & 53.93 &57.86 \\
\cmidrule{2-6}
&SOOD\dag~(\textbf{ours}) & - &\textbf{48.63} & \textbf{55.58} & \textbf{59.23} \\

\bottomrule
\end{tabular}
\end{table*}

\subsection{Implementation Details}

Without loss of generality, we take FCOS~\cite{tian2019fcos} as the representative anchor-free detector, and adopt ResNet-50~\cite{he2016deep} with FPN~\cite{lin2017feature} as the backbone for all our experiments. Following the previous works~\cite{xia2018dota,han2021align,han2021redet}, we crop the original images into $1024\times1024$ patches with a stride of 824, that is, the pixel overlap between two adjacent patches is 200. We utilize asymmetric data augmentation for unlabeled data. Specifically, we use strong augmentation for the student model and weak augmentation for the teacher model. Random flipping is used for weak augmentation, while strong augmentation contains random flipping, color jittering, random grayscale, and random Gaussian blur. All models are trained for 180k iterations on 2 RTX3090 GPUs. With the SGD optimizer, the initial learning rate of 0.0025 is divided by 10 at 120k and 160k. The momentum and the weight decay are set to 0.9 and 0.0001, respectively. Each GPU takes 3 images as input, where the proportion between unlabeled and labeled data is set to 1:2. The pseudo-label sampling ratio is set as 0.25 by default. Following previous SSOD works~\cite{zhou2022dense,liu2021unbiased}, we use the ``burn-in" strategy to initialize the teacher model.

\begin{figure*}[t]
	\begin{center}
        \includegraphics[width=0.8 \linewidth]{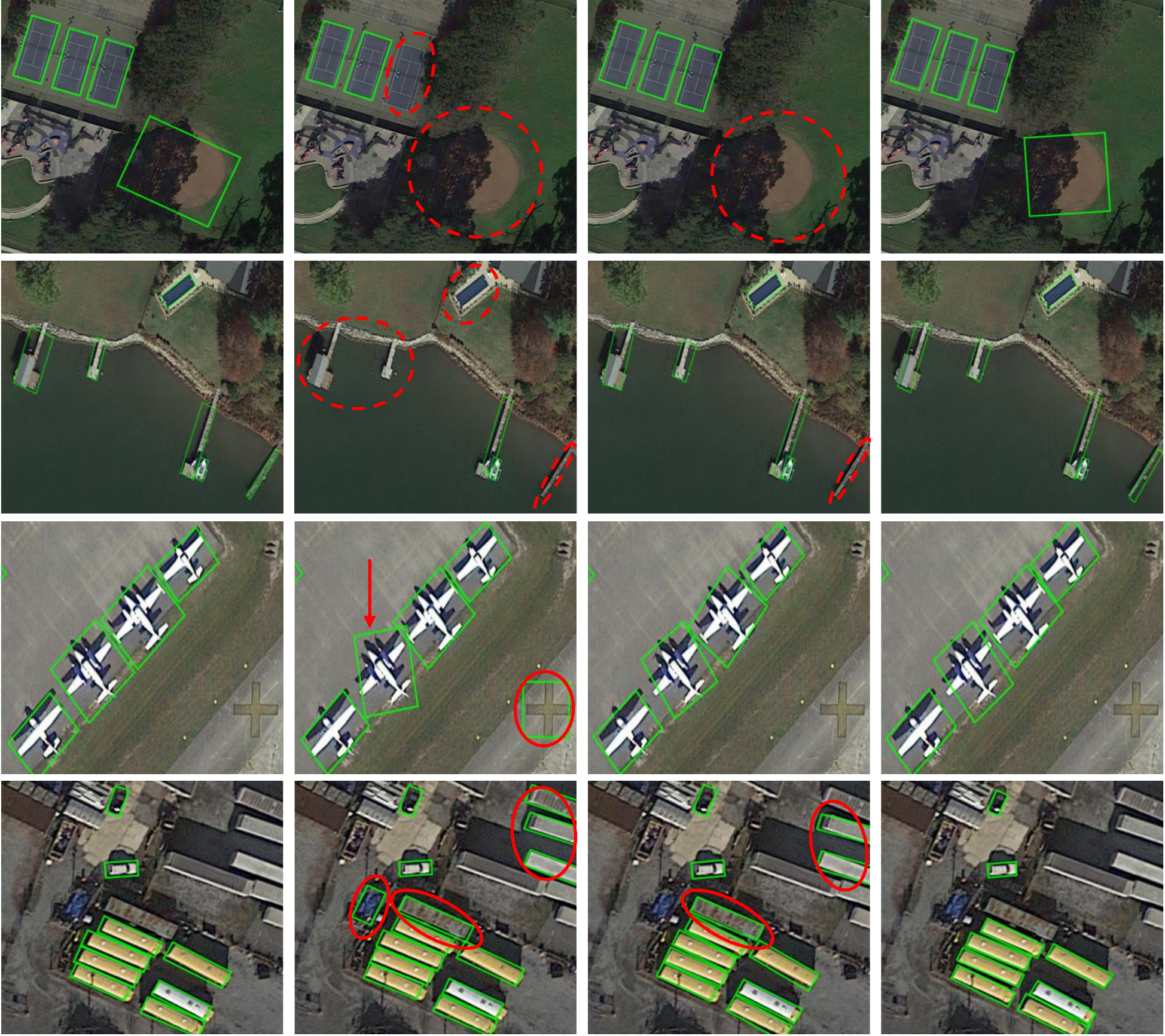}
	\end{center}
     \vspace{-10pt}
	\caption{Some visualization examples from the DOTA-v1.5 dataset. From left to right, each column shows ground truth, results of the supervised baseline (rotated-FCOS\cite{tian2019fcos}), Dense Teacher~\cite{chen2022dense}, and our SOOD. The green rectangles indicate predictions. The red dashed circle, solid red circle, and red arrow represent false negative, false positive, and inaccurate orientation prediction, respectively.}
   
	\label{fig:visualize}
\end{figure*}

\begin{table}[t]
\small
\centering
\setlength{\tabcolsep}{1.0mm}
\caption{Experimental results on DOTA-v1.5 under the Fully Labeled Data setting. * and \dag~indicate our implementations with rotated-Faster-RCNN and rotated-FCOS respectively. Numbers in font of the arrow indicate the supervised baseline.}
\label{tab:experiments_f}
\begin{tabular}{ccc}
\toprule
Method & Publication &mAP\\
\midrule
Unbiased Teacher*~\cite{liu2021unbiased} &ICLR 2021& 66.12 $\xrightarrow{-1.27}$ 64.85\\
Soft Teacher*~\cite{xu2021end}& ICCV 2021& 66.12 $\xrightarrow{+0.28}$ 66.40\\
Dense Teacher\dag~\cite{zhou2022dense}& ECCV 2022& 65.46 $\xrightarrow{+0.92}$ 66.38\\ 
\midrule
SOOD\dag~(\textbf{ours}) &- & 65.46 $\xrightarrow{+2.24}$ \textbf{67.70}\\

 \bottomrule
\end{tabular}
\end{table}

\subsection{Main Results}

In this section, we compare our method with the state-of-the-art SSOD methods~\cite{chen2022dense,xu2021end,liu2021unbiased} on DOTA-v1.5. For a fair comparison, we re-implement these methods on oriented object detectors with the same augmentation setting.

\vspace{0.5ex}\noindent\textbf{Partially Labeled Data.} We evaluate our method under different labeled data proportions, and the results are shown in Tab.~\ref{tab:experiments}. Our SOOD achieves state-of-the-art performance under all proportions. Specifically, it obtains 48.63, 55.58, and 59.23 mAP on 10\%, 20\%, and 30\% proportions, respectively, surpassing our supervised baseline by +5.85, +5.47, and +4.44 mAP. We also surpass the state-of-the-art anchor-free method Dense Teacher~\cite{zhou2022dense} by +1.73, +1.65, and +1.37 under various proportions. 
We provide two anchor-based methods for comparison, Unbiased Teacher~\cite{liu2021unbiased} and Soft Teacher~\cite{xu2021end}. On 10\% and 20\% proportions, our SOOD achieves higher performance than Soft Teacher, even though our baseline is weaker than Soft Teacher's. Under 30\% data proportion, our SOOD surpasses Soft Teacher and Unbiased Teacher by at least 1.40 mAP.

The qualitative results of our method compared with supervised baseline and Dense Teacher~\cite{zhou2022dense} are shown in Fig.~\ref{fig:visualize}. With the help of our RAW and GC, SOOD is able to exploit more potential semantic information from the unlabeled data, helping reduce false predictions and improve the detection quality.

\vspace{0.5ex}\noindent\textbf{Fully Labeled Data.} We also compare our SOOD with the other SSOD methods~\cite{zhou2022dense,xu2021end,liu2021unbiased} on fully labeled data setting. Since the reported methods are based on different detectors, we report the results of the compared methods and their baseline in Tab.~\ref{tab:experiments_f}. 

Our SOOD surpasses previous methods by at least 1.30 points. Compared to our baseline, we obtain +2.24 mAP improvement, which further demonstrates our method's ability to learn from unlabeled data. We notice that the performance of Unbiased Teacher~\cite{liu2021unbiased} drops after adding unlabeled data. The reason might be that Unbiased Teacher does not apply unsupervised losses for bounding box regression, which is important for oriented object detection.

\begin{table}[t]
\small
\centering
\setlength{\tabcolsep}{3.0mm}
\caption{The effectiveness of SOOD on other methods under Fully Labeled Data setting. * means based on RetinaNet~\cite{lin2017focal}. SOOD is able to generalize to other detectors and boost their performance.}
\label{tab:ablation_detectors}
\begin{tabular}{cccc}
\toprule
Detector & Publication & Method & mAP \\
\midrule
\multirow{2}{*}{CFA~\cite{guo2021beyond}} & \multirow{2}{*}{CVPR 2021} & Supervised & 65.75  \\
 & & Ours & \textbf{\vspace{1ex}67.07} \\
\midrule
\multirow{2}{*}{KLD*~\cite{yang2021learning}} & \multirow{2}{*}{NeurIPS 2021} & Supervised & 62.21 \\
 & & Ours & \textbf{64.62}  \\

\bottomrule
\end{tabular}
\end{table}

\noindent\textbf{Generalization on other detectors.}

To further validate the effectiveness of our method, we evaluate our method on other oriented object detectors, CFA~\cite{guo2021beyond} and KLD~\cite{yang2021learning}, under the Fully Labeled Data setting. As shown in Tab.~\ref{tab:ablation_detectors}, although CFA is a strong detector, our method still results in an improvement of +1.32 mAP and reaches 67.07 mAP. On the KLD detector, our method brings an improvement of +2.41 mAP. The above results validate the generalization ability of our method.

\subsection{Ablation Study}
In this section, we conduct extensive studies to validate our key designs. Unless specified, all the ablation experiments are performed using 10\% of labeled data. 

\vspace{1ex}\noindent\textbf{The effect of each component.}
We study the effects of the proposed two losses, Rotation-aware Adaptive Weighting~(RAW) loss and Global Consistency~(GC) loss. Note that our SOOD degrades to the vanilla dense pseudo-labeling framework without these two losses. As shown in Tab.~\ref{tab:ablation_components_10p}, both losses are proved effective and complementary under all three settings: RAW and GC can each bring performance gain, and the baseline is further improved when equipped with two losses. It indicates that the local constraint built by RAW and the global constraint built by GC can benefit the semi-supervised learning process, boosting the model by constructing one-to-one and many-to-many relationships between the teacher and the student.


\begin{table}[t]
\centering
\small
\setlength{\tabcolsep}{3.0mm}
\setlength{\abovecaptionskip}{0.5em}
\caption{The effects of Rotation-aware Adaptive Weighting (RAW) loss and Global Consistency (GC) loss. Experiments are conducted on 10\%, 20\% and 30\% labeled data settings.}
\label{tab:ablation_components_10p}
\begin{tabular}{cccccc}

\toprule
\multirow{2.5}{*}{Setting} & \multirow{2.5}{*}{RAW} & \multirow{2.5}{*}{GC} & \multicolumn{3}{c}{mAP} \\ 
\cmidrule{4-6} &&&10\% & 20\% & 30\% \\
\midrule
\uppercase\expandafter{\romannumeral1} & - & - & 47.24 & 54.07 & 57.74 \\
\uppercase\expandafter{\romannumeral2} & \checkmark & - & 47.82 & 55.21 & 58.93 \\
\uppercase\expandafter{\romannumeral3} & - & \checkmark & 47.71 & 54.72 & 58.70 \\
\uppercase\expandafter{\romannumeral4} & \checkmark & \checkmark & \textbf{48.63} & \textbf{55.58} & \textbf{59.23} \\
\bottomrule
\end{tabular}
\end{table}


\vspace{1ex}\noindent\textbf{The influence of sampling ratio.}
In this part, we discuss the influence of the ratios in sampling pseudo-labels. The results with different sampling ratios are shown in Tab.~\ref{tab:ablation_sample_ratio}. The best performance, 48.36 mAP, is achieved when the sample ratio is set to 0.25. Setting it to other values degrades the performance. We hypothesize that this value ensures a good balance between noises (e.g., false positives) and valid predictions (e.g., true positives). Increasing it will introduce more noise that harms the training process, while decreasing it leads to information loss and failure in learning the representation of objects.

\begin{table}[t]
\small
\centering
\setlength{\tabcolsep}{6.0mm}
\caption{The effect of the sampling ratio for instance-level dense pseudo-labeling. Experiments are conducted at 10\% setting, and the method is equipped with both RAW and GC losses.}
\label{tab:ablation_sample_ratio}
\begin{tabular}{ccc}
\toprule
Setting & Sample Ratio  & mAP\\
\midrule
\uppercase\expandafter{\romannumeral1} & 0.125 & 48.27  \\
\uppercase\expandafter{\romannumeral2} & 0.25 & \textbf{48.63} \\
\uppercase\expandafter{\romannumeral3} & 0.5 & 47.91 \\
\uppercase\expandafter{\romannumeral4} & 1.0 & 47.69\\

\bottomrule
\end{tabular}

\end{table}

\vspace{1ex}\noindent\textbf{The effect of different compositions in cost map of GC.}
Here, we study the effects of the spatial distance and the score difference when constructing the cost map of optimal transport in GC loss. The results of different settings are shown in Tab.~\ref{tab:ablation_components}. We get at most +0.28 mAP improvement when using only one of them, indicating that the information from only one side is inadequate for learning the global prior. When considering both the score difference and spatial distance, the performance gain brought by GC is further improved to +0.81 mAP. It indicates that the information from score difference and spatial distance are complementary. With their help, RAW can effectively model the many-to-many relationship between the teacher and the student, providing an informative guide to the model.

\begin{table}[t]
\centering
\small
\setlength{\tabcolsep}{6.0mm}
\caption{The effects of different compositions in the cost map of GC. Experiments are conducted at 10\% setting, and the method is equipped with RAW loss. The results indicate that both distance and score are essential factors of the cost map.}
\label{tab:ablation_components}
\begin{tabular}{cccc}
\toprule
Setting & Distance & Score & mAP\\
\midrule
\uppercase\expandafter{\romannumeral1} & - & - & 47.82\\
\uppercase\expandafter{\romannumeral2} & - & \checkmark & 48.10 \\
\uppercase\expandafter{\romannumeral3} & \checkmark & - & 47.94 \\
\uppercase\expandafter{\romannumeral4} & \checkmark & \checkmark & \textbf{48.63} \\
\bottomrule
\end{tabular}
\end{table}

\vspace{1ex}\noindent\textbf{The effect of RAW's hyper-parameter $\alpha$}.
Here, we study the influence of the hyper-parameter $\alpha$ in RAW. As shown in Tab.~\ref{tab:ablation_hyper_beta}, we set $\alpha$ to 1.0 and get the performance of 47.77 mAP. As $\alpha$ increases, the performance of our method improves when $\alpha$ varies from 1 to 50. However, further increasing it to 100.0 slightly hurt the performance. Therefore, we set it to 50 by default. For this observation, we conjecture that increasing the weight $\alpha$ will enlarge the influence of orientation information, but also amplify the impact of teacher's inaccurate labels.

\begin{table}[t]
\centering
\small
\setlength{\tabcolsep}{6.0mm}
\caption{The effect of hyper-parameter $\alpha$ in the RAW loss. Experiments are conducted at 10\% setting, and the method is equipped with GC loss.}
\label{tab:ablation_hyper_beta}
\begin{tabular}{cccc}
\toprule
Setting & $\alpha$ & mAP\\
\midrule
\uppercase\expandafter{\romannumeral1} & 1 & 47.77\\
\uppercase\expandafter{\romannumeral2} & 10 & 47.87\\
\uppercase\expandafter{\romannumeral3} & 50 & \textbf{48.63}\\
\uppercase\expandafter{\romannumeral4} & 100 & 47.95\\
\bottomrule
\end{tabular}

\end{table}

\subsection{Limitation and Discussion}
Although our method achieves satisfactory results on semi-supervised oriented object detection, the usage of aerial objects' characteristics is limited. Apart from orientation and global layout, many other properties of aerial objects should be considered, e.g., scale variations and large aspect ratios. Apart from that, we separately consider orientation and global layout by constructing two different constraints, which can be integrated into one unified module to utilize both information simultaneously. We also find that oriented objects and even complex objects wildly appear in other tasks, such as 3D object detection and text detection, leaving much room for further exploration.

\section{Conclusion}
In this paper, we have presented an effective solution for semi-supervised oriented object detection, which is important but neglected. Focusing on oriented objects' characteristics in aerial scenes, we have designed two novel losses, rotation-aware adaptive weighting (RAW) loss and global consistency (GC) loss. The former considers the importance of rotation information for oriented objects, dynamically weighting each pseudo-label-prediction pair by their rotation difference. The latter introduces the global layout concept to SSOD, measuring the global similarity between the teacher and the student in a many-to-many manner. To validate the effectiveness of our method, we have conducted extensive experiments on the DOTA-v1.5 benchmark. Compared with state-of-the-art methods, SOOD achieves consistent performance improvement on partially and fully labeled data.

\textbf{Acknowledgements.} This work was supported by the National Science Fund for Distinguished Young Scholars of China (Grant No.62225603) and the Young Scientists Fund of the National Natural Science Foundation of China (Grant No.62206103). 

{\small
\bibliographystyle{ieee_fullname}
\bibliography{egbib}
}

\clearpage

\begin{appendices}
\setcounter{table}{0} 
\setcounter{figure}{0} 
\setcounter{equation}{0} 

\renewcommand{\thefootnote}{\arabic{footnote}}

\begin{figure*}[t]
    \centering
    \includegraphics[width=0.76\linewidth]{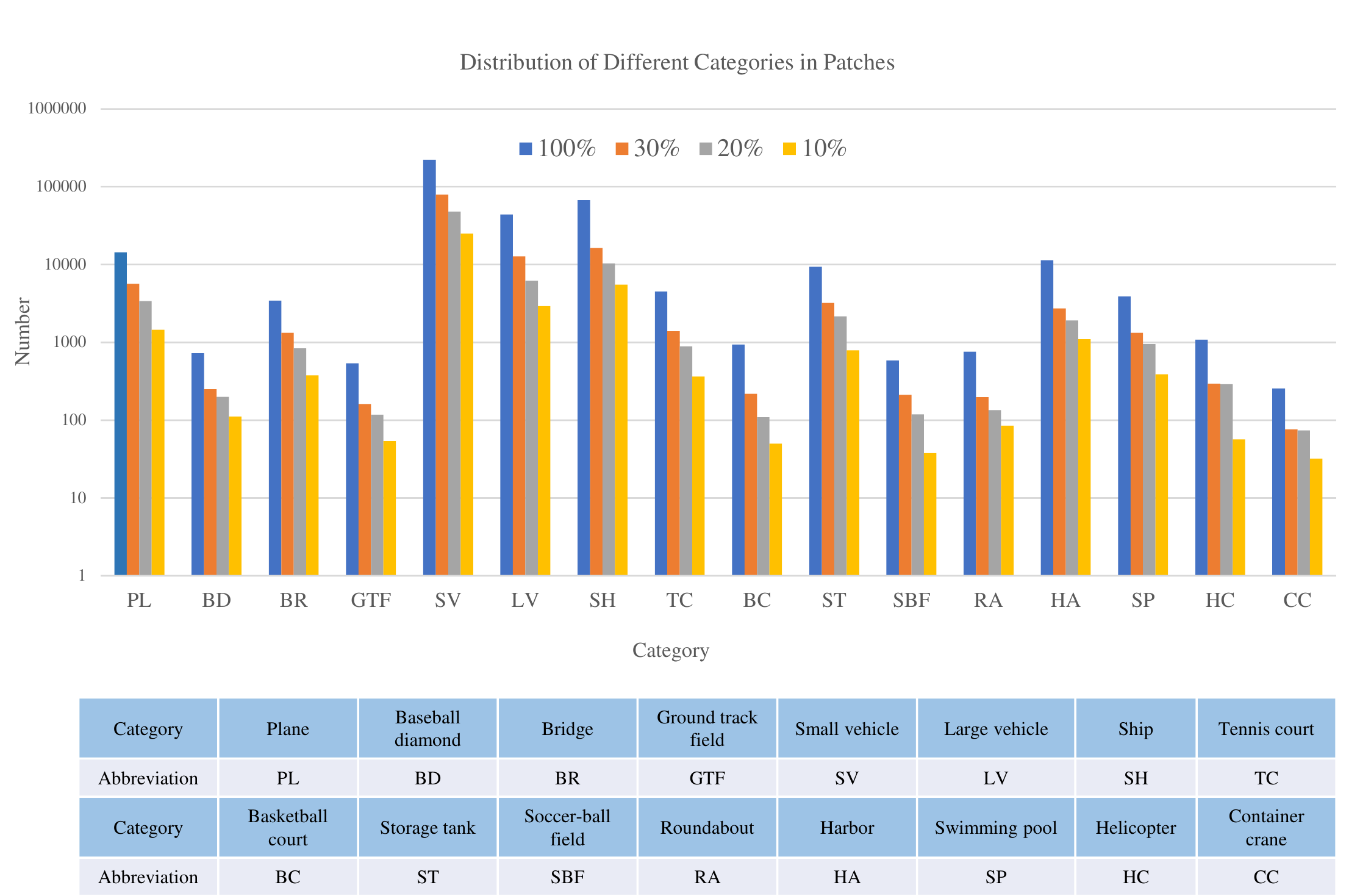}
    \caption{The top table shows distributions of different categories in the cropped image patches with abbreviations. The bottom table shows the corresponding abbreviation for each category. The distributions of different data splits are similar to the origin (100\%). In this case, these splits can well reflect the performance of different semi-supervised object detection methods.}
    \label{fig:appdenix_patch}
\end{figure*}

\begin{table*}[htbp]
\small
\centering
\setlength{\tabcolsep}{1.0mm}
\caption{Detailed comparison between SOOD and other methods. All experiments are conducted on DOTA-v1.5 under the Fully Labeled Data setting.}
\label{tab:detailed_expr}
\begin{tabular}{cccccccccccccccccc}
\toprule
Method  & PL & BD & BR &GTF &SV &LV & SH & TC & BC &ST & SBF & RA & HA & SP & HC & CC & mAP\\
\midrule
FCOS~\cite{tian2019fcos} & 89.8 &76.5 & 42.1& 56.9& 54.7 & 77.9 & 89.4 & 90.6 & 63.0 & \textbf{70.1} & 60.6& 66.0& 73.5& 65.5& 56.1& 14.7& 65.46\\
Dense Teacher~\cite{zhou2022dense} & \textbf{89.9} & 83.6 & \textbf{46.5} & 50.3& 53.6& 75.0& 88.7 & 90.7 & \textbf{69.5} & 67.9& 65.9& \textbf{73.3} & \textbf{73.7} & \textbf{67.5} & 56.9 & 9.1 & 66.38\\
\midrule
SOOD (ours) & 89.7 & \textbf{84.0} & 46.1 & \textbf{57.9} & \textbf{55.4} & \textbf{78.6} & \textbf{89.6} & \textbf{90.8} & 66.8 & 69.0& \textbf{69.5}& 70.1& \textbf{73.7} & 66.6 & \textbf{63.1} & \textbf{15.9} & \textbf{67.70}\\

\bottomrule
\end{tabular}
\end{table*}

\section{Details of Partially Labeled Data Sets}
We randomly sample 10\%, 20\%, and 30\% data from DOTA-v1.5-train~\cite{xia2018dota} to form partially labeled data sets. Besides, the 20\% set is a subset of the 30\% set, and the 10\% set is a subset of the 20\% set. To maintain the characteristic of the original data, we ensure the partially labeled sets have similar data distributions with DOTA-v1.5-train, as shown in Fig.~\ref{fig:appdenix_patch}. In this manner, these splits can well reflect the effectiveness of different semi-supervised object detection methods.

\section{Additional Experiments}

\subsection{Analysis on Result of Each Category}
As shown in Tab.~\ref{tab:detailed_expr}, we report the detailed results between our SOOD and other methods under the Fully Labeled Data setting. Our SOOD achieves 67.70 mAP, outperforming Dense Teacher~\cite{zhou2022dense} for most categories. Specifically, SOOD outperforms Dense Teacher by a large margin for categories named ground track field (GTF), helicopter (HC), and container crane (CC). We think the main reasons are two aspects: 1) the proposed rotation-aware adaptive weighting (RAW) loss can effectively utilize the orientation information, improving performance on orientation-sensitive objects like GTF and CC. 2) the proposed global consistency (GC) loss builds global constraint as an auxiliary, improving performance on dense objects like HC and CC. 

Although SOOD outperforms the supervised baseline on basketball court (BC) and roundabout (RA) by a lot, it is lower than Dense Teacher. It is likely that these two categories are often alone and similar to the background, which weakens the effect of GC and leads to sub-optimal supervision. Besides, even though SOOD performs better than Dense Teacher on storage tank (ST), it is worse than the supervised baseline. It may be because that RA is similar to the background and easily confused with other objects, resulting in too much noise in pseudo labels and worse performance.

\begin{table*}[htbp]
\small
\centering
\setlength{\tabcolsep}{4.0mm}
\caption{Comparison of our SOOD and other semi-supervised object detection methods using anchor-based detectors under the Fully Labeled Data setting. All methods are evaluated on DOTA-v1.5-val. * indicates implementation towards oriented objects.}
\label{tab:anchor_based_expr}
\begin{tabular}{ccccc}
\toprule
Detector & Method & Publication & mAP & $\Delta$ \\
\midrule
{\multirow{4}{*}{Faster R-CNN*~\cite{ren2015faster}}}&Supervised& NeurIPS 2016& 66.12 & - \\
& Unbiased Teacher~\cite{liu2021unbiased} & ICLR 2021 & 64.85 & -1.27 \\
& Soft Teacher~\cite{xu2021end} & ICCV 2021 & 66.40 & +0.28 \\
& SOOD & - & \textbf{66.64} & +0.52 \\
\midrule
{\multirow{2}{*}{Oriented R-CNN~\cite{xie2021oriented}}} & Supervised & ICCV 2021 & 67.26 & - \\
& SOOD & - & \textbf{68.50} & +1.24\\

\bottomrule
\end{tabular}
\end{table*}

\begin{table*}[ht]
\centering
\small
\setlength{\tabcolsep}{4.0mm}
\caption{Performance of SOOD on detector with stronger augmentation (e.g., multi-scale augmentation). The evaluation on DOTA-v1.5-test are conducted at public server\protect\footnotemark[1]{}, all methods are trained on DOTA-v1.5-train and DOTA-v1.5-val.}
\label{tab:strong_aug_expr}
\begin{tabular}{cccc}
\toprule
Detector & Method & DOTA-v1.5-val & DOTA-v1.5-test \\
\midrule
{\multirow{2}{*}{Oriented R-CNN~\cite{xie2021oriented} w/ multi-scale}} & Supervised & 68.55 & 75.67 \\
& SOOD & \textbf{71.04} & \textbf{76.42} \\

\bottomrule
\end{tabular}
\end{table*}

\subsection{SOOD with Anchor-Based Detectors}

We additionally adopt our SOOD on anchor-based detectors, e.g., rotated-Faster-RCNN~\cite{ren2015faster} and Oriented R-CNN~\cite{xie2021oriented}. As shown in Tab.~\ref{tab:anchor_based_expr}, our SOOD surpasses previous semi-supervised object detection methods on rotated-Faster-RCNN. Besides, on the state-of-the-art method Oriented R-CNN, SOOD still improves the performance by +1.24, which proves our SOOD is also suitable for anchor-based detectors.

\subsection{Impact of SOOD on Stronger Detector}

We adopt stronger augmentation on Oriented R-CNN~\cite{xie2021oriented} and evaluate the effectiveness of our SOOD, as shown in Tab.~\ref{tab:strong_aug_expr}. Note that for evaluation on DOTA-v1.5-test, we adopt both DOTA-v1.5-train and DOTA-v1.5-val for training as in~\cite{xie2021oriented}, using additional images from DOTA-v2.0~\cite{ding2021object} to form unlabeled data set\protect\footnotemark[2]{}. On DOTA-v1.5-val, SOOD improves the supervised baseline by +2.49 and reaches 71.04 mAP. Although with more labeled data for training, our SOOD can still boost the performance of the supervised baseline on DOTA-v1.5-test\footnotetext[1]{\url{https://captain-whu.github.io/DOTA/evaluation.html}}.

\begin{table}[!t]
\centering
\small
\setlength{\tabcolsep}{4.0mm}
\caption{The performance of GC on Soft Teacher.}
\label{tab:gc+soft_v2}
\begin{tabular}{cccc}
\toprule
\multirow{2.5}{*}{Setting} & \multicolumn{3}{c}{mAP} \\ 
\cmidrule{2-4}  &10\% &20\% & 30\% \\
\midrule
Soft Teacher & 48.46 & 54.89 & 57.83 \\
Soft Teacher + GC (\textbf{ours}) & \textbf{49.30} & 55.45 & 58.16 \\
SOOD (\textbf{ours}) & 48.63 &  \textbf{55.58} & \textbf{59.23}\\
\bottomrule
\end{tabular}
\end{table}

\subsection{Global Consistency Loss on Soft Teacher}

We additionally adopt Global Consistency (GC) loss on Soft Teacher~\cite{xu2021end} to evaluate its generalizability. As shown in Tab.~\ref{tab:gc+soft_v2},  GC improves Soft Teacher’s performance under three settings and surpasses SOOD under the 10\% setting. It indicates that GC can be easily applied to other semi-supervised paradigms.

\section{Impact of Global Consistency Loss}

\begin{figure*}[ht]
    \centering
    \includegraphics[width=1.0\linewidth]{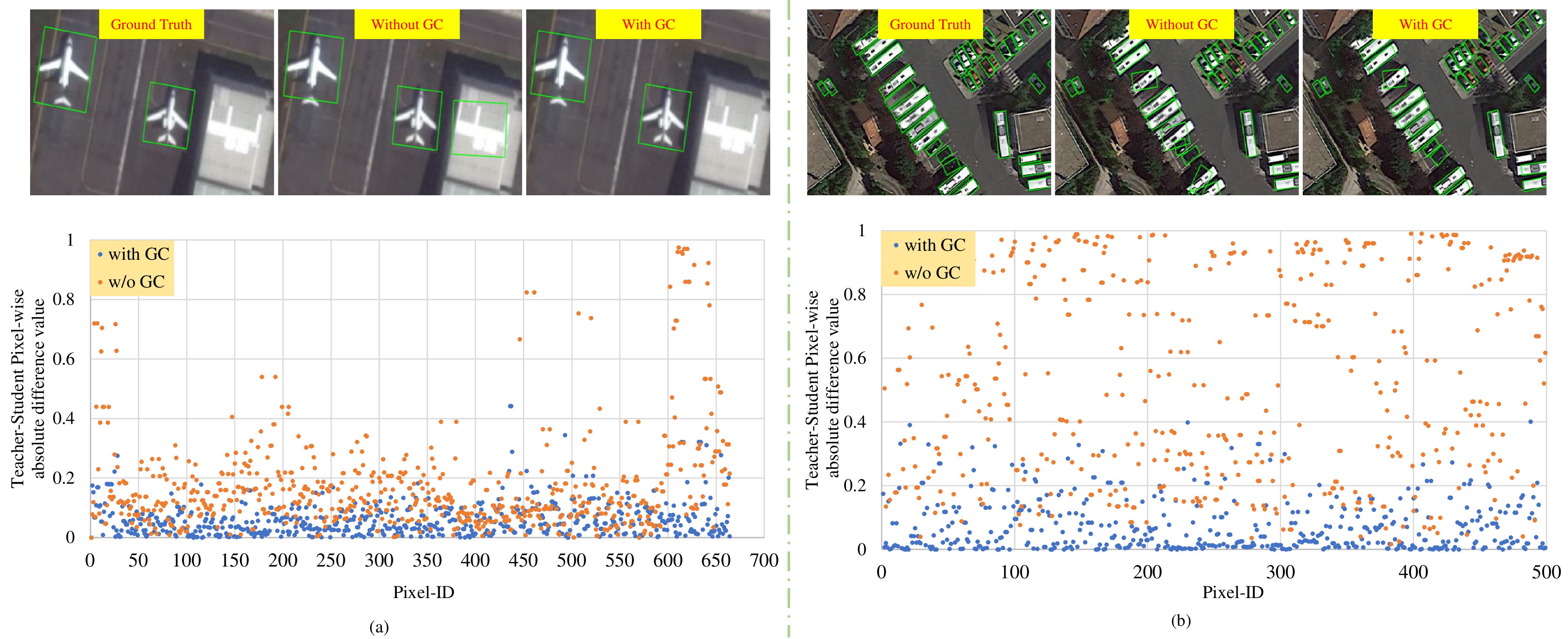}
    \caption{Visualization of the proposed GC's effect on prediction results and the distribution of absolute difference values between the teacher and student's classification scores. (a) represents sparse instances, and (b) represents dense instances. For the distribution map below, lower values indicate that the teacher and the student have more consistent distributions.}
    \label{fig:appdenix_gc_1}
\end{figure*}

We further provide qualitative visualizations to analyze the effect of the proposed Global Consistency (GC) loss. Specifically, we visualize some detection results of SOOD and SOOD without GC in Fig.~\ref{fig:appdenix_gc_1}, along with the distribution of absolute difference values between the teacher and the student's predictions. For the distribution map, lower values indicate that the teacher and the student's distributions are more consistent. Fig.~\ref{fig:appdenix_gc_1} shows that for both sparse and dense objects, GC can improve the consistency between the teacher and the student, leading to better prediction results.

\footnotetext[2]{We exclude the overlapped part between DOTA-v1.5 and DOTA-v2.0, using the left data of DOTA-v2.0 to form the unlabeled set.}

\newpage

\end{appendices}
\end{document}